\pdfoutput=1

\documentclass[11pt]{article}

\usepackage[]{acl}
\usepackage{booktabs, multirow} 
\usepackage{soul}
\usepackage{color}

\usepackage{times}
\usepackage{latexsym}
\usepackage{float}

\usepackage[T1]{fontenc}

\usepackage[utf8]{inputenc}

\usepackage{microtype}
\usepackage{multirow}
\usepackage{graphicx}
\usepackage{tabularx}
\title{UMASS\_BioNLP at MEDIQA-Chat 2023: 

Can LLMs generate high-quality synthetic note-oriented doctor-patient conversations?}

\author{Junda Wang \thanks{* indicates equal contribution}  \, \, Zonghai Yao \footnotemark[1] \, \,  Avijit Mitra \, \, Samuel Osebe \, \, Zhichao Yang \, \, Hong Yu \\
\\
\bf{CICS, University of Massachusetts, Amherst, MA, USA} \\ 
\\
\bf{jundawang@umass.edu  \, \, zonghaiyao@umass.edu} \\
}

\begin{document}
\maketitle
\begin{abstract}

This paper presents UMASS\_BioNLP team participation in the MEDIQA-Chat 2023 shared task for Task-A and Task-C. We focus especially on Task-C and propose a novel LLMs cooperation system named a doctor-patient loop to generate high-quality conversation data sets. The experiment results demonstrate that our approaches yield reasonable performance as evaluated by automatic metrics such as ROUGE, medical concept recall, BLEU, and Self-BLEU. 
Furthermore, we conducted a comparative analysis between our proposed method and ChatGPT and GPT-4. This analysis also investigates the potential of utilizing cooperation LLMs to generate high-quality datasets. \footnote{Our codes are released at \url{https://github.com/believewhat/Dr.NoteAid}}

\end{abstract}

\section{Introduction}

The issue of the growing burden of clinical documentation has become a critical concern in healthcare, resulting in increased job dissatisfaction and burnout rates among clinicians and adversely affecting patient experiences. Nevertheless, timely and accurate documentation of patient encounters is crucial for safe, effective care and communication between specialists. Consequently, there is a growing interest in automating assisting doctors in diagnosis based on Large Language Models (LLMs) due to its remarkable advancement in the field of artificial intelligence (AI), being highly sophisticated systems that have been extensively trained on massive amounts of textual data.~\cite{brown2020language,Sanh2021,chowdhery2022palm,Longpre2023TheFC,openai2023gpt4}

The swift progress of AI and its extensive influence on various fields have garnered considerable attention from the research community. One notable area is the creation of instruction-following LLMs~\cite{touvron2023llama, alpaca, vicuna2023, zhu2022minigpt4}, which demonstrate extraordinary ability in understanding instructions and producing human-like responses. These auto-regressive LLMs undergo a two-step process: they are initially pre-trained on web-scale natural languages through next-token prediction and subsequently fine-tuned to comply with extensive human instructions~\cite{dale2021gpt}. This method leads to impressive performances across a broad range of natural language processing (NLP) tasks and generalizes to unseen tasks, underscoring their potential as a comprehensive solution for diverse challenges, including natural language understanding, text generation, and conversational AI~\cite{floridi2020gpt}. Many auto-regressive LLMs, such as ChatGPT, have further training with RLHF to align with human preference and finally allow these models to generate content that most people prefer. In the biomedical domain, many researchers have attempted to apply auto-regressive models to medical tasks such as patient triage~\cite{Levine2023TheDA}, automatic disease coding~\cite{Yang2022MultilabelFI}, and doctor-chatbot~\cite{yunxiang2023chatdoctor, xu2023baize}.

However high-quality dialogue datasets featuring doctor-patient interactions are a task that is inherently complex. One major difficulty in constructing such a dataset is the sensitive nature of the content, as healthcare conversations often involve private and confidential patient information~\cite{kelly2019key,rindfleisch1997privacy, annas2003hipaa}. Ensuring privacy protection and adhering to strict data regulations, such as HIPAA, becomes crucial in the development process. Consequently, the compilation of authentic doctor-patient dialogues requires careful consideration of privacy and data protection measures to prevent potential ethical and legal concerns. 
Recent work~\cite{mts-dialog, aci-demo} attempts to synthesize data by letting humans play the roles of doctor and patient in a conversation, but the huge cost makes the research community seek the help of LLMs or Chatbot models to simulate such role-playing game (doctor and patient) for data augmentation.
However, recent Chatbot models~\cite{yunxiang2023chatdoctor, zeng2020meddialog} are only based on single-turn or multi-turn question-and-answer repositories rather than real conversations between patients and doctors. Question-and-answer datasets lack logical coherence, whereas real conversations can help the model understand the proper order of questions~\cite{drew2001conversation}, thereby guiding patients to describe their symptoms accordingly and ultimately aiding in disease diagnosis.
So we cannot simply use the recent medical chatbot models to generate high-quality note-oriented doctor-patient conversations.
On the other hand, the success of these LLM models remains heavily reliant on human input to guide conversations in the right direction. This dependence necessitates users to provide relevant and precise prompts based on their intentions and the chat agent's feedback, which can be challenging, time-consuming, and occasionally unfeasible. In healthcare contexts, individuals without medical expertise may struggle to generate appropriate prompts for directing communicative agents to deliver accurate medical advice or diagnoses~\cite{tang2023does, Liao2023DifferentiateCA}. 

All of this 
raises a crucial question:
How can we try role-playing games to guide conversations toward clinical note completion in healthcare settings without a large number of human annotations?
To address these issues, 
we propose a cutting-edge cooperative agent framework, Doctor-Patient Loop.
This approach involves two main ChatGPT agents taking on the roles of doctor and patient in dialogue, with additional ChatGPT agents responsible for fact-checking, ensuring conversations remain focused on the provided notes, determining when the dialogue should be terminated, and refining the conversation to enhance its coherence and fluency. The collaboration among multiple ChatGPT agents leads to the creation of more realistic doctor-patient dialogue datasets, which in turn can be utilized for training models that better mimic genuine healthcare communication scenarios.

In this paper, we conducted a series of experiments with the help of the data set of the MEDIQA-Chat competition shared task. Specially, we present our entry for Task-A and Task-C. We explored a new approach to solve For Task-A. We trained BioMedLM \footnote{https://github.com/stanford-crfm/BioMedLM} on the dataset of Task-A and designed prompts for different section headers. For Task-c, we explored the potential for creating scalable methods that promote autonomous cooperation among communicative agents in medical settings. We construct a doctor-patient loop to generate high-quality clinical dialogue. Our paper's contributions can be summarized as follows:

\begin{itemize}
    \item[1)] We propose innovative approaches to foster autonomous cooperation among communicative agents in medical settings, highlighting their cognitive processes and collaborative capabilities.
    \item[2)] We concentrate on the generation and utilization of continuous doctor-patient dialogue datasets, which serve as valuable resources for developing AI systems that can better understand and address context-sensitive inquiries in healthcare communication.
    \item[3)] We finetuned BioMedLM on a collection of data sources to obtain the FLAN-BioMedLM model and then finetuned this model on the Task-A dataset on the classification task. It achieved good performance in the task of section header classification and was used to assist ChatGPT in generating clinical notes.
\end{itemize}

\section{Related Work}
The MEDIQA-Chat 2023 tasks \footnote{https://sites.google.com/view/mediqa2023}~\cite{abacha2023empirical, mediqa-chat-2023, mts-dialog,aci-demo} focused on both Dialogue2Note Summarization and Note2Dialogue Generation tasks. The researchers constructed a novel dataset comprising 1,700 doctor-patient conversations (16k turns and 18k sentences) and their summarized clinical notes (6k sentences). They also proposed an investigation of standard evaluation metrics, domain-specific metrics, and expert judgments for the task, including the calculation of the correlation between the automatic and manual scores for the evaluation of the generated clinical notes. In this paper, we use a cooperative agent framework to generate the conversation data sets.

For the \textbf{section header and content classification} (Task-A), the SOAP (Subjective, Objective, Assessment, and Plan) structure is commonly used by providers \cite{podder2021soap}. 
The Subjective section is a detailed report of the patient’s current conditions, such as source, onset, and duration of symptoms, mainly based on the patient’s self-report. This section usually includes a history of present illness and symptoms, current medications, and allergies. The Objective section documents the results of physical exam findings, laboratory data, vital signs, and descriptions of imaging results. The Assessment section typically contains medical diagnoses and reasons that lead to medical diagnoses. The assessment is typically based on the content from the subjective and objective sections. The Plan section addresses treatment plans based on the assessment. Previous work focused on identifying these four general SOAP sections \citep{Kwon2022AnAS}. In this paper, we focused on predicting the specific subsections.

For the \textbf{Dialogue2Note Summarization} task (Task-A\&B), there are many solutions already in the industry~\cite{krishna2021extracting, song-summarizing, yim2021towards, krishna2021extracting, schloss2020towards}. This process generally follows a similar pipeline. Clinical conversations are initially recorded and then transcribed. Subsequently, the utterances are classified into various medical sections, and clusters of utterances containing medically relevant information for each section are predicted from the transcript. Finally, a section-conditioned summarization model is employed to generate concise summaries for each cluster of utterances associated with their respective sections. However, the size of their private training data is  larger than the training data of this competition. Thus, we applied LLM for this competition. Instruction finetuning helps LLM generalize into unseen tasks where training data is limited \citep{Longpre2023TheFC}. For example, an instruction-finetuned 11B-param model outperforms the 60B-param model without instruction-finetuning in the BIG-Bench dataset \citep{Chung2022ScalingIL}. Thus, we instructed finetuned BioMedLM model for Task-A Dialogue2Note Summarization task.

For the \textbf{Note2Dialogue Generation} task (Task-C), MEDIQA-Chat 2023 treats it as a data augmentation task. Recent investigations into utilizing LLMs for data augmentation have produced notable results. \citet{li2023feasibility} explored the possibility of using LLMs to generate training data for tasks such as code summarization, code translation, and code generation. In a similar vein, \citet{Dai2023ChatAugLC} suggested employing LLMs to tackle low-resource scenario model training by augmenting data to enhance performance. Moreover, \citet{gilardi2023chatgpt} and \citet{ding2022gpt} studied the effectiveness and accuracy of LLMs for data annotation, respectively, achieving promising outcomes even when compared with Crowd-Workers. \citet{bonifacio2022inpars} utilized LLMs to create positive sample pairs for training downstream models. At the same time, \citet{zhou2022large} focused on generating appropriate prompts with LLMs to improve performance further. Lastly, \citet{dai2022promptagator} mainly targeted few-shot retrieval tasks, combining LLMs with a limited number of samples to produce additional training data for retrieval models. In the biomedical field, \citet{tang2023does} investigated the potential of LLMs in clinical text mining and introduced a novel training paradigm to address suboptimal performance and privacy concerns. \citet{Liao2023DifferentiateCA} examined responsible and ethical Artificial Intelligence Generated Content (AIGC) in medicine, analyzing differences between human-authored and LLM-generated medical texts and developing machine learning workflows for efficient detection and differentiation. In this paper, we explored our cooperative agent framework's performance in Task-C.

\section{Methods}

\subsection{MEDIQA-Chat Tasks}
The competition proposed two new shared tasks, namely Dialogue2Note and Note2Dialogue, which aim to facilitate clinical note creation through the summarization of medical conversations and the generation of synthetic doctor-patient dialogues for data augmentation purposes, respectively.
\begin{itemize}
    \item[1)] Dialogue2Note Summarization: This task entails generating a clinical note that succinctly summarizes a conversation between a doctor and a patient. The resulting clinical note may contain one or multiple sections, such as Assessment, Past Medical History, and Past Surgical History. Task-A focuses on generating specific note sections from the doctor-patient conversation: first, predicting the section heading and then generating the content of the specific section.
    \item[2)] Note2Dialogue Generation: This task involves creating a synthetic doctor-patient conversation based on the information provided in a full clinical note. Participants are required to generate a dialogue that effectively captures the context and content of the original clinical note, thereby contributing to relevant data creation and augmentation.
\end{itemize}
These shared tasks, as presented at the ACL conference, are expected to significantly contribute to the development of cutting-edge methodologies and techniques in the realm of automatic clinical note generation, ultimately improving the overall efficiency and quality of healthcare documentation.

\subsection{Conversation to Clinical Note}
Similar to the general training pipeline of FLAN, we first instruction-finetuned BioMedLM (2.7 billion parameters model pre-trained on PubMed articles) on a collection of data sources to obtain the FLAN-BioMedLM model and then finetuned this model on Task-A dataset. Our approach differed from FLAN in the specific model and the data collection tailored for Task-A.
Since this is a medical-domain conversation task, we selected related data sources including 12 medical-domain generation datasets and 4 general-domain conversation/controlled generation datasets as shown in table \ref{tab: flan_tasks}. Medical-domain generation tasks include long-form question answering between doctors and patients, squad-like question answering from medical notes, medical notes summarization, research article summarization, and abbreviation disambiguation.
This task collection contains 110 prompt templates and 400 million tokens. 

\begin{table}
\centering
\scriptsize
\resizebox{0.7\linewidth}{!}{
\begin{tabular}{lrl}\toprule
Domain &Dataset &Reference \\\midrule
\multirow{12}{*}{Medical} &MeQSum &\citep{ben-abacha-demner-fushman-2019-summarization} \\
&Primock57 &\citep{papadopoulos-korfiatis-etal-2022-primock57} \\
&EmrQA &\citep{pampari-etal-2018-emrqa} \\
&DiSCQ &\citep{lehman-etal-2022-learning} \\
&MEDIQA-AnS &\citep{Savery2020QuestiondrivenSO} \\
&Pubmed-ccdv &\citep{cohan-etal-2018-discourse} \\
&Medal &\citep{wen-etal-2020-medal} \\
&Diagnoise-me &\citep{Zeng2020MedDialogLM} \\
&Medmcqa &\citep{Pal2022MedMCQAA} \\
&Ebm\_pico &\citep{nye-etal-2018-corpus} \\
&Pubhealth &\citep{kotonya-toni-2020-explainable-automated} \\
&Pmc\_patients &\citep{Zhao2022PMCPatientsAL} \\\midrule
\multirow{4}{*}{General} &Multiwoz &\citep{zang-etal-2020-multiwoz} \\
&Taskmasters &\citep{byrne-etal-2019-taskmaster} \\
&Dart &\citep{nan-etal-2021-dart} \\
&WebNLG &\citep{shimorina-gardent-2018-handling} \\
\bottomrule
\end{tabular}
}
\caption{Datasets used to train FLAN-BioMedLM}
\label{tab: flan_tasks}
\vspace{-4mm}
\end{table}

We then finetuned FLAN-BioMedLM on Task-A. Specifically, we built a pipeline to classify section heading first, and then used this heading to generate section content. When the generated heading string did not match to the ground truth class name, we used fuzzy string matching to find its nearest valid header. We finetuned FLAN-BioMedLM on these two subtasks separately. Our prompts are shown in table~\ref{tab: prompttaska}.

\begin{table*}
\begin{tabular}{ll}
\hline
\multirow{7}{*}{\textbf{Heading subtask}} & Dialogue: \textcolor{blue}{dialogue}. \\& Given the dialogue above, select a section of the medical note from \\& the options below. \\& Options: history of present illness; review of systems; past medical history; \\& medications; chief complaint; past surgical history; disposition; \\& diagnosis; emergency department course; plan; labs; assessment; \\& allergy; gynecologic history; exam; other history; procedures; imaging; \\& immunizations; family history social history.\\
\hline
\multirow{2}{*}{\textbf{Content subtask}} & Dialogue: \textcolor{blue}{dialogue}. \\& Generate section \textcolor{red}{heading}  of the medical note from dialogue.\\
\hline

\end{tabular}
\caption{FLAN-BioMedLM prompt template for Task-A, where colored words will be replaced with actual content.}
\label{tab: prompttaska}
\end{table*}

\begin{table*}
\centering
\begin{tabular}{ccccccccc}
\hline
Model & R-1 & R-2  & R-L & R-L-Sum & bertscore\_f1	& bleurt \\
\hline
FLAN-BioMedLM & 0.3283	& 0.1351	& 0.2743	& 0.2743	& 0.6699	& 0.4757\\
ChatGPT & 0.3828	& 0.1828	& 0.3158	& 0.3166	& 0.7015	& 0.5405\\
\hline
\end{tabular}
\caption{Synthetic data quality evaluation on Task-A. }
\label{taska}
\vspace{-4mm}
\end{table*}


We also explored this task using ChatGPT. We found that ChatGPT has a lower accuracy in classifying section headings, and its performance in generating notes is highly dependent on the given examples. Therefore, we first use FLAN-BioMedLM for headings classification and then provide ChatGPT with corresponding examples based on the section headings.

\subsection{Clinical Note to Conversation}

\subsubsection{Segmentation}
In MEDIQA-Chat Task-C, the training set consists of comprehensive and extensive clinical notes. There are 20 validation samples and 40 test samples. We try to apply GPT3.5-turbo to generate the dialogue. However, due to the maximum token limitation imposed by the GPT3.5-turbo API, it is infeasible to input the entire dialogue when providing a prompt. Consequently, we dissect the clinical note into several section headings as shown in the heading subtask of Table~\ref{tab: prompttaska}. For each section heading, we leverage the dataset from Task-A to construct a prompt that assists the model in generating a dialogue segment. Ultimately, the conversation fragments corresponding to different section headings are concatenated to form a complete dialogue.
    
\subsubsection{Doctor-Patient Loop}
Language models often lack sufficient medical knowledge to help them accomplish the target tasks~\cite{sung2021can, yao2022extracting, yao2022context}. So we employed the MedSpaCy library to extract relevant CUI codes from clinical notes, aiming to guide subsequent conversations around these key terms. Such a checklist can help our pipeline improve factuality~\cite{tang2022understanding, abacha2023investigation, chang2023revisiting}, and can be changed very flexibly for other purposes, like information retrieval~\cite{khattab2022demonstrate}, entity linking~\cite{yao2020zero}, medical jargon extraction~\cite{kwon2022medjex}, causulity~\cite{yuan2023causality}, and rules or knowledge injection~\cite{fei2021enriching, yao2021improving, 10.1093/jamia/ocad081}.
Upon extraction, we initiated a doctor-patient loop involving multiple rounds of dialogue to generate comprehensive conversations. In each round, one ChatGPT instance played the role of a doctor while the other acted as a patient. The doctor, incorporating the case details and identified keywords, would select and focus on up to four key terms to pose questions to the patient. The ChatGPT representing the patient would then respond to the inquiries based on the clinical notes. In each dialogue round, the conversation history from previous rounds was incorporated as a prompt input to ensure context continuity and coherence throughout the interaction, finally completing a loop in the process. Furthermore, we also construct a factuality-checking module to ensure the comprehensiveness of our conversation. We employed a ChatGPT-based approach to monitoring whether the dialogue encompasses all essential information.

\begin{table*}
\centering
\begin{tabular}{l|cccccccc}
\hline
 &\textbf{R-1} & \textbf{R-2} & \textbf{R-L} & \textbf{R-L-Sum} &  \textbf{C-R}  & \textbf{BLEU}& \textbf{SBLEU $\downarrow$}  &\textbf{Len}\\
\hline
& \multicolumn{8}{c}{MEDIQA-CHAT-2023-RESULTS Task-C Test dataset}\\
\hline
1. Cadence     &54.36	& 23.81	& 20.64	& 47.45 & - & -& -&-\\
2. UMass\_BioNLP &42.36	&11.96	&15.96	&40.46 &- & -&-&- \\
3. NUSIDS     &40.63	& 14.18	& 17.24	& 39.45 & - & -& -&-\\
\hline
& \multicolumn{8}{c}{Additional Experiments and Results \textcolor{red}{(1. done after competition 2. Eval dataset)}}\\
\hline
ChatGPT-short     & 48.31 & 17.43 & 19.33 & 50.74 & 35.42 & 4.00 & 0.018 & 46.5\\
GPT-4-short       & 53.16 & \textbf{19.49} & \textbf{23.10} & 50.39 & 44.95 & 6.13 &0.016 & 42.0\\
Ours-short & \textbf{54.18} & 17.43 & 19.33 & \textbf{50.74} & \textbf{47.19} & \textbf{6.62}& \textbf{0.013}  & 45.1 \\
\hline
ChatGPT-long     & 48.56 & 16.74 & 22.41 & 46.36 & 35.75 & 4.93& 0.017 & 62.8\\
GPT-4-long       & 53.29 & 20.20 & \textbf{24.06} & 50.81 & 45.69 & 5.92& 0.019 & 58.1\\
Ours-long       & \textbf{56.48} & 19.74 & 20.03 & \textbf{53.41} & \textbf{51.23} & \textbf{6.12}& \textbf{0.017}  & 62.5 \\
\hline
\end{tabular}
\caption{Synthetic data quality evaluation on MEDIQA-Chat using auto-metrics.}
\label{results}
\vspace{-4mm}
\end{table*}

\subsection{Evaluation}
All methods are evaluated with Rouge-1, Rouge-2, Rouge-L, Rouge-Lsum \cite{lin2004rouge} on both Task-A and Task-C. For Task-A, we also apply BERTScore-F1~\cite{zhang2019bertscore} and BLEURT~\cite{sellam2020bleurt} to test our result. For Task-C, we also use BLEU score~\cite{post2018call}. 
For Task-C, to measure the generated text including all the important information such as symptoms or medication from the clinical domain, we used another metric named Concept-Recall, which evaluates the overlap of clinical keywords present in the two texts. We first extracted all Unified Medical Language System (UMLS)~\cite{bodenreider2004unified} entities from text using MedSpaCy~\cite{eyre2021launching}. We further refined this list of entities by selecting only those that were clinically important. Specifically, we included entities whose semantic groups are diseases, drugs, devices, and procedures as defined in~\cite{bodenreider2003exploring}, and exclude other semantic types such as fish, bird, and other conceptual entities. Finally, we calculated the overlap of entities from generated text and reference text by recall scores. 
We also evaluated the text diversity in Task-C. \citet{zhu2018texygen} proposed a benchmarking platform for text generation models that is fully open-sourced. We followed their work and evaluated the diversity of the generated conversation based on their proposed Self-BLEU score.

\section{Experiment}
In this section, we discuss our proposed methods' performance on MEDIQA-Chat 2023 Task-A and Task-C.
All the detailed experiment settings can be found on our GitHub.

\subsection{Task-A} 
We compared FLAN-BioMedLM and ChatGPT in the two subtasks.
In the heading classification task, FLAN-BioMedLM achieved an accuracy of 0.705, and ChatGPT scored an accuracy of 0.355.
However, ChatGPT outperformed FLAN-BioMedLM in the content generation task, as shown in Table ~\ref{taska}. 

\subsection{Task-C}
As of the end of the competition, the results of our method in the competition ROUGE family are shown in Table \ref{results}. After the end of the competition, we further did follow-up prompt engineering and saw a significant improvement in the results. In this and the next section, our discussions are all based on new results. In order to be fair, we can't compare the new results with other teams in the competition, so our baseline is mainly ChatGPT and GPT4. 

In our study, we found that ChatGPT and GPT-4 are highly sensitive to the choice of the prompt. To achieve optimal performance, we experimented with various prompts and categorized them into two groups: one for generating short-length conversations with an average length of around 40 utterances and the other for generating long conversations exceeding 50 utterances. We also observed that the length of the conversation has a significant impact on the Rouge score and concept score, as shown in the table~\ref{results}. 

Due to the API's maximum token limit, ChatGPT and our method (based on ChatGPT) could not generate long conversations. We found that the length of the conversation has a significant impact on the score, and scores tend to improve when the conversation length approaches that of human conversations. Therefore, we optimized the combined prompt to only concatenate the next conversation segment with the one generated from the previous topic. This allowed us to generate longer conversations within the maximum token limit. As a result, our Rouge total scores have further improved. 

In addition, we found that ChatGPT and GPT-4 are suitable for generating conversations of moderate length. When we forced them to generate very long conversations, GPT4 will generate highly repetitive sentences and diverge significantly from real conversations. ChatGPT will divide long utterances into several short utterances. Hence, both ChatGPT and GPT-4 struggle to cover all the essential information even if we force them to generate longer conversations, and their concept recall scores were lower than our model's. Even in their longer versions of conversations, the amount of information covered was less than that of our shorter version because in the experiment result our shorter version model's concept is 47.19 indicating that our model can include most information and the Self-BLEU score is 0.013 which demonstrate the diversity of our model. For the longer version, our model sacrificed a small amount of diversity but gained a significant improvement in concept recall (51.23) and Rouge score. Therefore, the experiment result can demonstrate that segmentation can guide ChatGPT to cover all the essential medical information. In the segmentation module, we provide separate prompts for each different section header to guide the model's attention to the corresponding important information. Furthermore, the doctor-patient loop can make the generated conversations more logical, and the maximum turn setting ensures that the model covers all the key phrases.

\subsection{Case Study}

In this section, we provide examples of conversations in Table~\ref{case1} generated by our model and some prompts (Table~\ref{prompt}) to demonstrate that our approach can produce more human-like conversations. Our system mainly consists of the following prompts:

\paragraph{Doctor Prompt} is utilized to instruct the model to assume the role of a physician, asking logically coherent questions based on the patient's clinical note and previous dialogue for the purpose of generating dialogue datasets.

\paragraph{Patient Prompt} is designed to guide the model to play the role of a patient, answering the doctor's questions based on their own medical history. We set the patient's level of education to be low to ensure that ChatGPT's language style is more similar to that of an actual patient in daily conversation.

\begin{table}[H]
\begin{tabularx}{0.48\textwidth}{p{7.4cm}} 
\hline
\small{\textbf{Doctor Prompt}} \\ 
\hline
\small{Clinical Note: \textcolor{blue}{Note}}\\
\small{Please role-play as a doctor and further ask a question based on the above dialogue to follow up the history conversation. The treatment plan, medication, and dosage you give to the patient must also be consistent with the clinical note. Your question should be around these keywords, and you cannot modify these keywords or use synonyms.}\\
\small{Key Words:\textcolor{red}{$key_{1},key_{2},...$}}\\
\hline

\small{\textbf{Patient Prompt}}\\ 
\hline
\small{Clinical Note: \textcolor{blue}{Note}}\\
\small{Please act as a patient and answer my question or follow up on the conversation. Your answer must be consistent with the clinical note and cannot include information that is not in the clinical note. Your responses should be more colloquial.}\\
\hline

\small{\textbf{Polish Prompt}} \\
\hline
\small{Please rewrite all the conversations based on the notes to become fluence and more colloquial, like a normal conversation between the doctor and patient based on the clinical notes. Now you should rewrite the following conversations, and your conversation should include all the information and all the keywords. The keywords must be used directly instead of using synonyms when using them in the conversation} \\
\small{Key Words: \textcolor{red}{$key_{1},key_{2},...$}} \\
\small{The conversation:" \textcolor{purple}{Conversation}} \\
\small{Clinical Note: \textcolor{blue}{Note}} \\
\small{The conversation between the doctor and the patient should involve multiple rounds, with each question and answer being relatively short. You should try to ensure that the dialogue is smooth.} \\
\hline

\small{\textbf{Hallucination Prompt}} \\
\hline
\small{Check whether the information of the conversation is consistent with the clinical note. If there is some information that you cannot find on the clinical note, please eliminate it. You also should delete the duplicate part. The conversation should include all the key words:\textcolor{red}{$key_{1},key_{2},...$}}\\
\small{Clinical Note: \textcolor{blue}{Note}} \\
\small{Conversation: \textcolor{purple}{Note}}\\

\hline

\small{\textbf{Postediting Prompt}}\\
\hline
\small{The above two paragraphs were extracted from a complete conversation. Please concatenate the two dialogues together. It means that your generation should include all the information such as the dosage of the medication  which is mentioned in the clinical note. You should try to ensure that the dialogue is smooth. The conversation must include these key words:\textcolor{red}{$key_{1},key_{2},...$} and you should also eliminate the repeat parts.}\\
\small{History Conversation: \textcolor{purple}{Conversation 1}} \\ 
\small{Generated Conversation: \textcolor{brown}{Conversation 2}} \\ 

\hline
\end{tabularx}
\caption{Prompts for Task-C.}
\label{prompt}
\vspace{-4mm}
\end{table}

\paragraph{Polish Prompt} is utilized to modify previously generated conversations, enhancing their coherence and naturalness to resemble real-life conversations between doctors and patients. It also ensures that no important information is omitted during the conversation.

\paragraph{Hallucination Prompt} is used to remove information generated by the model that is inconsistent with the clinical note. This ensures that the content generated by the model is more aligned with the clinical note.

\paragraph{Postediting Prompt} is utilized to merge conversations from two different sections into a single smooth dialogue, ensuring that the conversation occurs at the same time and place.

By comparing our model to the conversations generated by GPT-4 (Table~\ref{case2}), we observed that our generated conversations are more logical and resemble human-to-human conversations more closely. For instance, in the GPT-4 generated conversation, the doctor did not ask any questions regarding symptoms such as fevers, chills, dizziness, nausea, vomiting, or diarrhea, as highlighted in the red utterance, but the patient answered that they did not have those symptoms. Regarding the conversations generated by ChatGPT (Table~\ref{case3}), we observed that the sentence structure of the questions posed by the doctor was highly uniform and the patient responses were quite mechanical, failing to resemble human-to-human conversations. Although GPT-4 somewhat addresses this issue, it still exhibits similar limitations. By contrast, our proposed method generates non-repetitive questions and replies, resulting in a more realistic dialogue dataset.

\section{Conclusion}

In conclusion, our work proposes innovative approaches to promote autonomous cooperation among communicative agents in medical settings to generate the doctor-patient dialogue datasets as valuable resources for developing AI systems that can better understand and address context-sensitive inquiries in healthcare communication. To validate the quality of our generated datasets, we establish a hybrid evaluation system consists of Rouge scores, concept recall, Sarebleurt and Self-bleurt. We use our evaluation system to compare our model with GPT4 and ChatGPT. We demonstrate that our method contributes to advancing the development of AI systems in healthcare communication and paves the way for more effective doctor-patient interactions.

\section{Limitation}

Although our method delivers optimal results, it doesn't comply with data protection regulations like HIPAA, even though Azure offers a HIPAA-compliant option. From a privacy standpoint, deploying a local model such as LED might be preferable, but our findings indicate that further work is needed for this method to achieve satisfactory performance. Regardless, when creating automated conversation-generation systems, healthcare providers and developers must ensure that the entire system—including text-to-dialogue, data transmission and storage, and model inference—complies with privacy and security standards to maintain trust and avoid privacy breaches in clinical environments. Hence, developing an automated conversation generation system from clinical note entails several ethical considerations.

Firstly, obtaining informed consent is vital: patients should be informed about their recordings, and data ownership must be emphasized. Equitable access is also crucial; the system must accommodate patients from various backgrounds, including those with disabilities, limited technical literacy, or language barriers. Lastly, continuous evaluations are required to ensure that the system's performance does not deteriorate and adversely affect the quality of care.

\begin{table}
\begin{tabularx}{0.48\textwidth}{p{7.5cm}}
\hline
\textbf{ChatGPT} \\
\hline
\tiny{\textbf{Doctor:} Good morning, Brian. How have you been feeling lately?}\\
\tiny{\textbf{Patient:} I have been feeling out of sorts lately, not sure if it is due to the change in seasons or due to performing lots of projects and some construction on my home.}\\
\tiny{\textbf{Doctor:} That sounds exhausting. Have you been experiencing any symptoms?}\\
\tiny{\textbf{Patient:} Yes, I have been feeling fatigued and lightheaded for about five weeks now.}\\
\tiny{\textbf{Doctor:} \textcolor{blue}{Have you noticed any shortness of breath or chest discomfort while exerting energy?}}\\
\tiny{\textbf{Patient:} Yes, I have experienced some shortness of breath and chest cramps.}\\
\tiny{\textbf{Doctor:} \textcolor{blue}{Have you been coughing lately?}}\\
\tiny{\textbf{Patient:} Yes, I have a slight cough, but I am not sure if it's due to the change in seasons.}\\
\tiny{\textbf{Doctor:} \textcolor{purple}{How has your diet been lately?}}\\
\tiny{\textbf{Patient:} My diet has been a little bit of a struggle. We have been eating less healthy food due to the construction work in the kitchen.}\\
\tiny{\textbf{Doctor:} \textcolor{purple}{How have you been doing with your salt intake?}}\\
\tiny{\textbf{Patient:} I have been pretty good with my salt intake.}\\
\tiny{\textbf{Doctor:} \textcolor{purple}{How has your heart failure been?}}
\tiny{\textbf{Patient:} I have been pretty good with my diet since last year and have been staying on top of that as much as possible. I have continued to utilize Lasix daily.}\\
\tiny{\textbf{Doctor:} \textcolor{purple}{How has your hypertension been?}}\\
\tiny{\textbf{Patient:} It has been well controlled with lisinopril 20 mg a day. I have been monitoring my blood pressure regularly.}\\

\tiny{...}\\

\hline
\end{tabularx}
\caption{ChatGPT data sample for Task-C.}
\label{case3}
\end{table}

\begin{table}
\begin{tabularx}{0.48\textwidth}{p{7.5cm}}
\hline
\textbf{Our:}\\
 \hline
\tiny{\textbf{Doctor:} Good morning, Brian. How are you feeling today?}\\
\tiny{\textbf{\textbf{Patient:} } Good morning, Doctor. I'm feeling a little out of sorts lately. I'm not sure if it's because of the change in seasons or because I've been doing a lot of projects around the house.}\\
\tiny{\textbf{\textbf{Patient:} } Sure. I've been feeling fatigued and lightheaded for about five weeks now. And when I'm active, I've been experiencing some shortness of breath and chest cramps. I've also had a slight cough, but I'm not sure if it's just because of the change in seasons.}\\
\tiny{\textbf{Doctor:} Have you noticed any bloating or changes in your diet?}\\
\tiny{\textbf{\textbf{Patient:} } Yeah, I've been feeling bloated every once in a while. And my diet has been a little bit of a struggle because we've had construction going on in our kitchen, so we've been eating less healthy food.}\\
\tiny{\textbf{Doctor:} Okay, let's talk about your medical history. You have a history of congestive heart failure, correct?}\\
\tiny{\textbf{\textbf{Patient:} } Yes, that's correct.}\\
\tiny{\textbf{Doctor:} And how have you been managing that?}\\
\tiny{\textbf{\textbf{Patient:} } I've been pretty good with my salt intake, and I've been keeping up with my diet since the last year. And I've been taking Lasix daily.}\\
\tiny{\textbf{Doctor:} Okay, good to know that. And what about your hypertension?}\\
\tiny{\textbf{\textbf{Patient:} } That's been well controlled with lisinopril 20 mg a day. I've been monitoring my blood pressure regularly.}\\
\tiny{\textbf{Doctor:} Great. Did you experience any other symptoms that you haven't mentioned yet, such as swelling in the lower extremities, fever, chills, dizziness, nausea, vomiting, weight gain, or diarrhea?}\\
\tiny{\textbf{\textbf{Patient:} } No, I didn't.}\\
\tiny{\textbf{Doctor:} Okay, thanks for letting me know. Based on what you've told me, I'm going to order some tests, and we'll follow up after those come back. In the meantime, try to get some rest and eat healthier if possible.}\\
\tiny{\textbf{\textbf{Patient:} } Alright, thank you.}\\
\tiny{...}\\

\hline
\end{tabularx}
\caption{Our pipeline sample for Task-C.}
\label{case1}
\end{table}

\begin{table}
\begin{tabularx}{0.48\textwidth}{p{7.5cm}}
\hline
\textbf{GPT4} \\
\hline
\tiny{\textbf{Doctor:} Good morning, Mr. White. How have you been feeling since your last visit?}\\
\tiny{\textbf{Patient:} Good morning, doctor. I've been feeling a bit out of sorts lately. I'm not sure if it's due to the change in seasons or because I've been doing a lot of projects and construction on my home.}
\tiny{\textbf{Doctor:} Can you be more specific about how you've been feeling?}\\
\tiny{\textbf{Patient:} Sure, I've been experiencing fatigue and lightheadedness for about 5 weeks now. When I exert energy, I sometimes get short of breath and have chest cramps.}\\
\tiny{\textbf{Doctor:} That's concerning. \textcolor{blue}{Have you noticed any other symptoms, like a cough or fever?}}\\
\tiny{\textbf{Patient:} I've had a slight cough, but I'm not sure if it's just the change in seasons. No fever though.}\\
\tiny{\textbf{Doctor:} \textcolor{blue}{Have you experienced any changes in your diet or eating habits recently?}}\\
\tiny{\textbf{Patient:} Yes, actually. We started doing construction on our kitchen over Labor Day weekend, and since then we've been eating less healthy food.}\\
\tiny{\textbf{Doctor:} And how have you been managing your heart failure and hypertension? \textcolor{blue}{Have you been monitoring your salt intake and taking your medications regularly?}}\\
\tiny{\textbf{Patient:} I've been pretty good with my salt intake and have been watching my diet for the past year. I take Lasix daily for my heart failure, and my hypertension is controlled with lisinopril 20 mg a day. I monitor my blood pressure regularly.}\\
\tiny{\textbf{Doctor:} That's good to hear. You mentioned earlier that you filled out a review of the systems sheet when you checked in. Did you experience any weight gain, swelling in the lower extremities, or other symptoms?}\\
\tiny{\textbf{Patient:} \textcolor{red}{No, I didn't experience any weight gain or swelling in my lower extremities. Also, no fevers, chills, dizziness, nausea, vomiting, or diarrhea.}}\\
\tiny{\textbf{Doctor:} Alright, let's move on to the physical examination. I'm going to check your neck for any jugular venous distention.}\\
\tiny{\textbf{Patient:} Sure, go ahead.}\\

\tiny{...}\\

\hline
\end{tabularx}
\caption{GPT4 data sample for Task-C.}
\label{case2}
\end{table}

\newpage

\bibliography{clinical_nlp23} 
\bibliographystyle{acl_natbib}




\end{document}